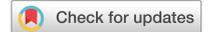

# Automated player identification and indexing using two-stage deep learning network

Hongshan Liu[1], Colin Adreon[2], Noah Wagnon[2], Abdul Latif Bamba[3], Xueshen Li[1], Huapu Liu[4], Steven MacCall[4] & Yu Gan[1]✉

American football games attract significant worldwide attention every year. Identifying players from videos in each play is also essential for the indexing of player participation. Processing football game video presents great challenges such as crowded settings, distorted objects, and imbalanced data for identifying players, especially jersey numbers. In this work, we propose a deep learning-based player tracking system to automatically track players and index their participation per play in American football games. It is a two-stage network design to highlight areas of interest and identify jersey number information with high accuracy. First, we utilize an object detection network, a detection transformer, to tackle the player detection problem in a crowded context. Second, we identify players using jersey number recognition with a secondary convolutional neural network, then synchronize it with a game clock subsystem. Finally, the system outputs a complete log in a database for play indexing. We demonstrate the effectiveness and reliability of player tracking system by analyzing the qualitative and quantitative results on football videos. The proposed system shows great potential for implementation in and analysis of football broadcast video.

American football is one of the most popular sports worldwide. The 2022 Super Bowl had 99.18 million television viewers in the US and 11.2 million streaming viewers around the world[1]. With more sophisticated developments in data analysis and storage, there has been significant growth in the demand for data-intensive analytic applications in American football.

Player identification is a standard way of interpreting sports videos and helping commentators explain games on television. Different from other sports, game interpreting and data analytics in American football are largely based on the concept of play. The plays of an American football game can be represented as a sequence of non-overlapping time segments of a running game clock that generally start with the movement of the ball from the center offensive lineman called the "snap", and end immediately at the time of each play's conclusion. American football leagues at several levels of competition such as the National Football League (NFL)[2] and the National Collegiate Athletics Association (NCAA)[3] publish statistical guidance manuals that fully and thoroughly document various types of plays so that when individual games are logged play-by-play, the statisticians and loggers are following standardized play definitions, including how to account for the game clock time that has elapsed during individual play occurrences so that there is no overlapping game clock data across two or more plays.

Player identification facilitates the comprehensive indexing of game videos based on player participation per play. However, in order to meet this demand, data analysis methods need to be specifically designed in comparison with traditional manual methods. Such traditional methods are time-consuming and impractical for handling a large number of football games recorded today. Recent advances in image processing techniques have opened the door for many interesting and effective solutions to this problem. Player identification can be classified into two main categories: Optical Character Recognition (OCR) based and Convolution Neural Networks (CNN) based approaches[4].

Before deep learning became popular, OCR-based approaches segmented images into HSV color space and classified numbers based on segmentation results[5]. In another example[6], bounding boxes of players were extracted by a deformable part model detector, then performed OCR and classification. However, deep learning models have been shown to provide greater capacity in object detection tasks and present more accurate results[7].

[1]Biomedical Engineering, Stevens Institute of Technology, Hoboken 07030, NJ, US. [2]Electrical and Computer Engineering, The University of Alabama, Tuscaloosa 35487, AL, US. [3]Electrical Engineering, Columbia University in the City of New York, New York, NY 10027, US. [4]Library and Information Science, The University of Alabama, Tuscaloosa, AL 35487, US. ✉email: ygan5@stevens.edu

The CNN approach in[8] detected players by the histogram of oriented gradients and a linear support vector machine (SVM) and classified player numbers within a bounding box by training a deep convolutional neural network. There are efforts[8–12] in exploring jersey number recognition by adopting various deep learning models. Gaussian Mixture Model (GMM) is utilized in[13] for background subtraction to locate moving foreground objects from the static background and to detect moving players from the field. In[14], mask R-CNN and YOLOv2 were compared for player detection using the pre-trained models because of a lack of annotated data. Liu et al.[15] proposed a player detection system in basketball using motion detection algorithms. Deep learning approach was used to analyze cricket batting[16]. Key methods for analyzing cricket batting are also explored[17] through key event detection and event importance scoring. Similarly, an SVM classifier is employed in a few Khan et al. papers[18,19] to detect scoreboxes in the video frames. The extracted features are used as input to the classifier, which learns to distinguish between scoreboxes and non-scorebox regions. Once a scorebox is detected, the method further localizes its position within the frame by refining the initial detection with contour analysis, template matching, or edge detection in order to accurately determine the precise boundaries of the scorebox. These localizations are also seen in Guo et al. Localization and Recognition paper[20], where they utilize Scale-Invariant Feature Transform (SIFT)-based feature extraction, point matching for scoreboard localization, team recognition techniques for extracting scoreboard information, and post-processing steps to achieve accurate localization and recognition of scoreboards in sports videos. Advancements in computer vision and pattern recognition allow for specific bodily poses to be processed in a much more efficient and accurate manner[21]. Pose estimation is widely utilized to guide the player detection deep learning network[22–25]. Recent effort[23] explored human body part cues and proposed a faster region-based convolution neural network (R-CNN) approach for player detection and number recognition, which works well when the player and number are shown clearly to camera but is not often the case for broadcast video. To address the player identification problems in broadcast video, Senocak et al.[22] utilized body part features to design a system to recognize body representation at multi-scale and with enhanced accuracy. An adaptive real-time player segmentation approach[26] was proposed by training a student network with the data annotated by a teacher network, which saves manual annotating. However, segmentation requires more computational resources, and the use of mask R-CNN causes degraded performance for complex scenarios.

Most of the existing sports analysis studies have focused on player detection in hockey[10], basketball[12,23] and soccer[8,27]. These studies are not focused on an in-depth analysis of football player detection. Football player identification is challenging due to the frequent clustering pattern of crowded players. Besides the exploration of player identification, specific player detection systems for football games have not been studied very extensively. To date, no database systems for automatic jersey number detection and player information indexing have been reported.

Although different player-detecting strategies have been explored in various sports, the achievement of fast and accurate player detection in American football remains elusive. A key element of a successful analysis system is detection accuracy. However, camera blur, player motion, and small targets pose many challenges to achieving satisfactory accuracy for player identification. In addition to player identification, complicated game clock management rules in American football also added challenges in indexing game logs.

In comparison with most sport analysis tasks, players participating in football games are usually crowded together and are prone to occlusion. The size of players also varies significantly depending on their positions relative to the camera. The small size of objects increases the difficulty of identifying players. In addition, the instantaneous and rapid movement of the players can blur the image and distort their shape. Different poses during falling and stumbling can further make detection more challenging. A motion-based background subtraction method performs weakly on blurred and occluded objects. Classical object detection algorithms give unsatisfactory results when dealing with crowded settings because overlapping objects can lead to false negative predictions. Since useful information is ignored, sports analysis systems bring incomplete results that may mislead viewers.

In this work, we develop an innovative football analysis system that automatically tracks players on a per-play basis and identifies any time the play clock is actuated during the game. The proposed system operates directly on the video footage and simultaneously extracts player information and game clock information on a frame-by-frame basis, while automatically generating logs for play indexing. Figure 1 shows the typical output of the proposed system.

We utilize a detection transformer-based network to detect players in a crowded setting and employ a focal loss-based network as a digit recognition stage to index each player given an imbalanced dataset. We then capture game clock data per play. The football analysis system outputs a database containing an index of each play in the game, as well as a list of quarters, game clock start/end times, and participating players. We summarize our contributions as:

1. We propose, to our best knowledge, the first deep learning-based football analysis system for jersey number identification and logging.
2. We devise a two-stage neural network to address player detection and jersey number recognition, respectively. The two-stage design highlights the player-related region for precise jersey number recognition. The first stage, a transformer design, addresses the issue of identifying a player in a crowded setting. The second stage further addresses the issue of data imbalance in jersey number identification.
3. We build a database for play indexing. The system automatically extracts game clock information along with identified jersey numbers, making it possible to generate a game log for player indexing.
4. We conduct experiments on American football videos. The mean average precision of our proposed system achieves 87.9% in player detection and 91.7% in jersey number recognition.

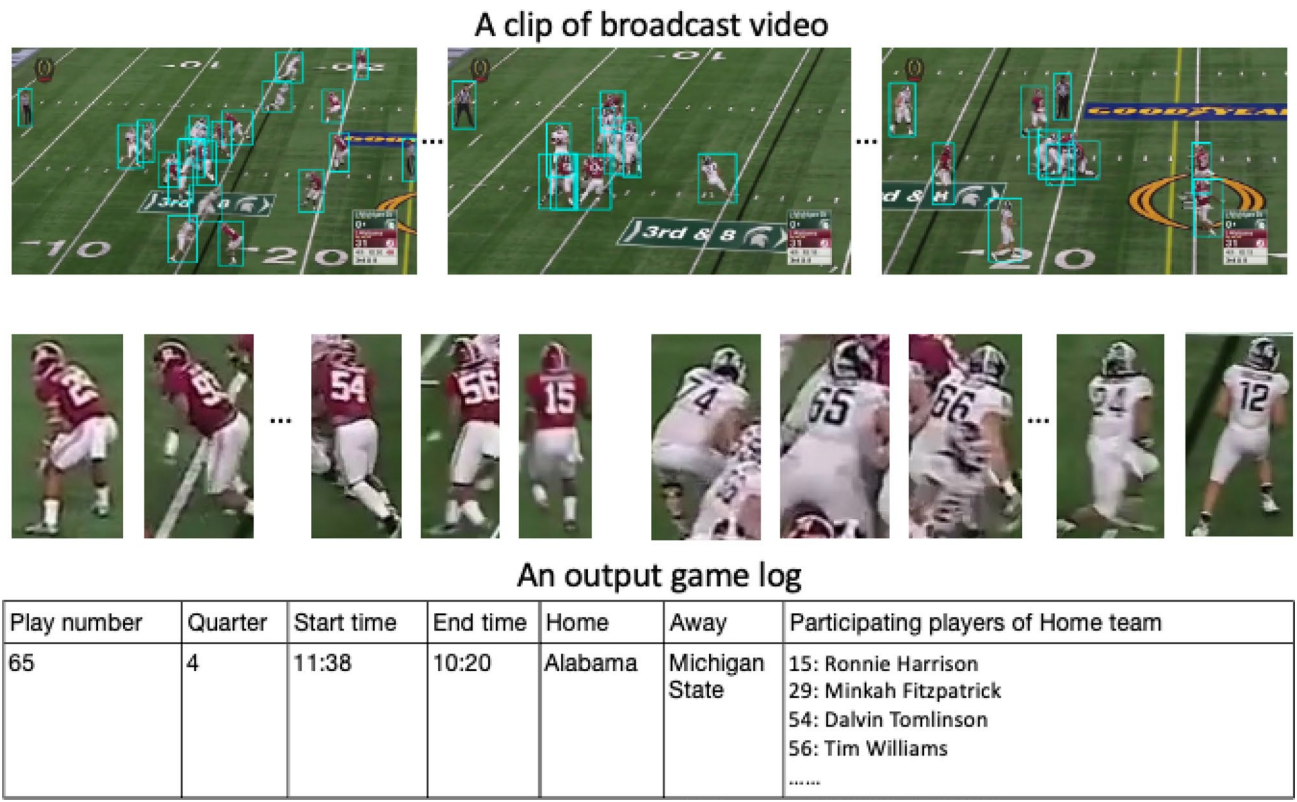

**Figure 1.** An example of the output of the proposed system: player detection, jersey number recognition, and game log.

## Methodology

**Overall framework.** We propose an automated football analysis system to index each play in a game by quarter, game clock start/end time, and a list of participating players, as shown in Fig. 2.

*Two-stage design.* We choose a two-stage object detection network to enhance the capability of small object-detection in high definition videos. In sports broadcasting and player identification, the input image is usually in high definition. Conventional object detection networks usually downsize an input image to a specific size, e.g. YOLO[28] resizes input to 416 × 416, SSD[29] resizes input to 512 × 512, which makes the player region unrecognizable. As shown in Fig. 3, a one-stage YOLOv2[30] leads to an unsatisfactory performance for two aspects. First, multiple players are identified within a single bounding box, lacking capability to differentiate crowded players. Second, jersey number information are eliminated in downsized image, as limited number of pixels associated with a jersey number. It thus is prone to detection error. To address this issue, we propose a two-stage design to first highlight a player region and then zoom into the player region for jersey number recognition.

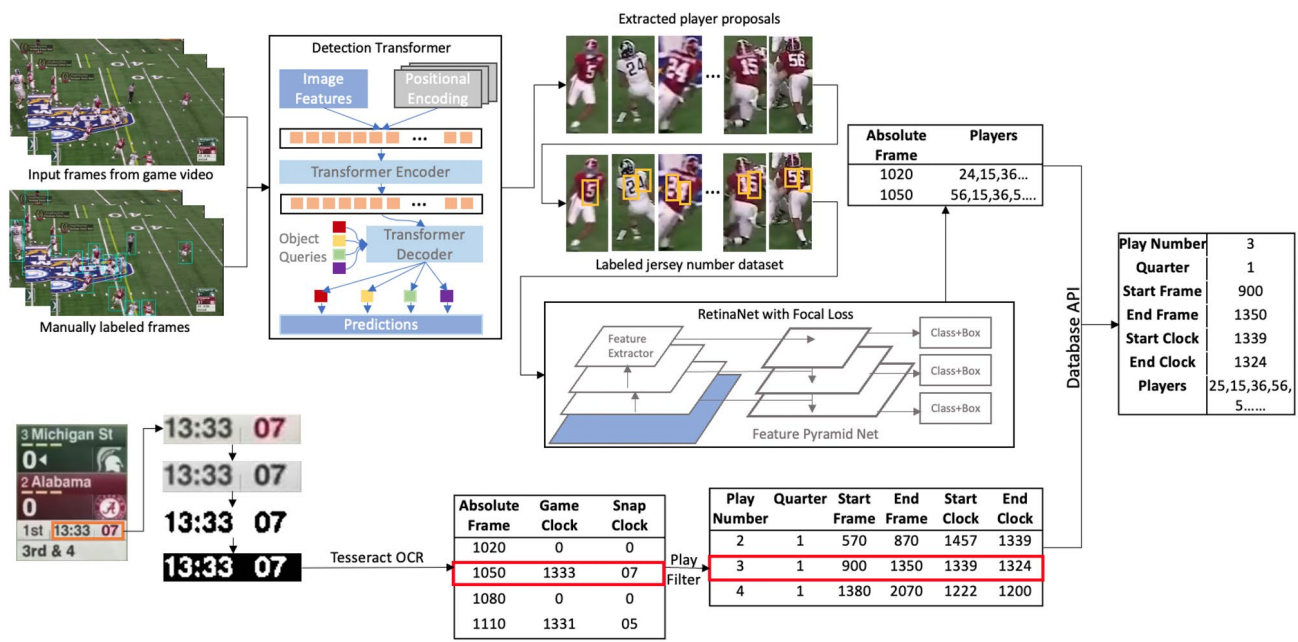

**Figure 2.** Flowchart of the proposed work.

*Three subsystems.* The overall flowchart consists of three subsystems of pattern recognition: player detection, jersey number recognition, and clock identification. For the first subsystem, we employ the Detection Transformer[31], which can detect players within extremely crowded scenario in order to address the technical challenge caused by highly overlapped targets. The second subsystem processes the detected player proposals and identifies jersey numbers accordingly. The digit recognition subsystem is implemented using RetinaNet50[32]. The third subsystem extracts game clock information for play indexing. Because of the presence of rewinds and timeouts in the video, the exact timestamp of the game needs to be extracted from each frame. The subsystem operates on frame-by-frame basis to extract game clock time using the Tesseract for optical character recognition. Clock information gets filtered into plays, which provides additional information of the estimated time of each play within a game. Upon completing the three subsystems, the information discovered in the jersey number is further associated to a readable and comprehensive database indexed by its game clock timestamp.

**Player detection subsystem.** Two main challenges are presented for player detection. First, we use data augmentation to alleviate data variation caused by motion effects, such as blurry and distorted objects. Second, we employ DEtection TRansformer (DETR)[31] to address the player detection problem against crowded background settings.

To ensure our method is robust to frames that are motion-corrupted, we augment our training dataset with more instances having motion effects. Specifically, we apply a Gaussian filter[33] to a portion of training images to create realistic motion-corrupted images. The training images are also scaled for data augmentation, as shown in Fig. 4.

Inspired by the transformer model in natural language processing, the player detection model is capable of identifying overlapped bounding boxes by reasoning their spatial relationships of those bounding boxes. Combining a set-based Hungarian loss which enables unique matching between predictions and ground-truth, DETR solves the set prediction problem with a transformer. Compared to conventional object detection methods, such as Faster R-CNN[34], DETR benefits from the bipartite matching and attention mechanism of the transformer encode-decode architecture[35], and therefore it excels in solving the detection task within a crowded setting of a football game.

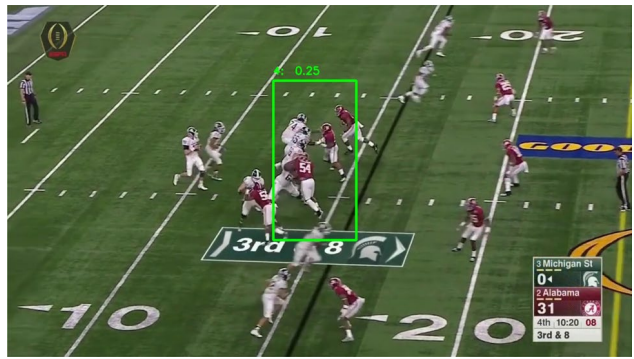

**Figure 3.** Image processed with one-stage YOLOv2, shows limited performance if a one-stage model is directly applied.

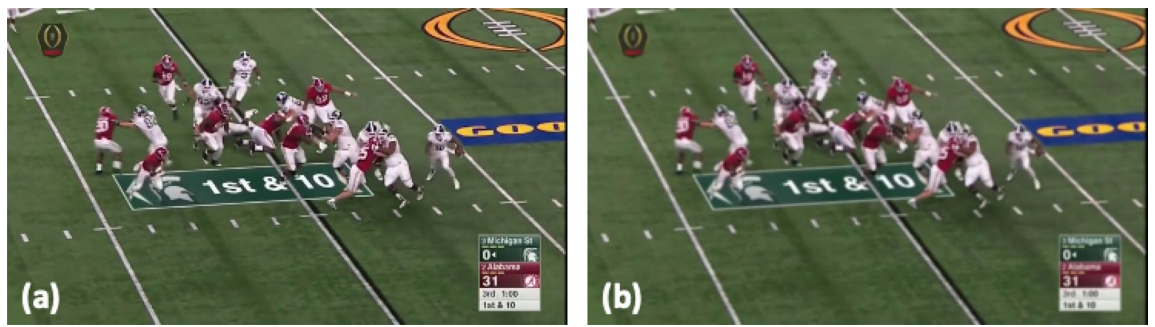

**Figure 4.** Example of data augmentation: (**a**) Original frame; (**b**) Original frame with motion-blurry effect.

The DETR framework in our implementation stems from ResNet50[36], a conventional CNN backbone. The feature map is extracted from the input image, then augmented with a positional encoding and then fed into the transformer encoder. Each encoder layer has a multi-head self-attention module to explore the correlation within the input of each layer. For the transformer decoder, the learnable object queries pass through each decoder layer with a self-attention module to explore the relations within itself. In the decoder, another cross-attention module takes the key elements of the feature map from the encoder output as side input to let object queries cross-learn the features from the feature map. At the last step, a feed-forward network (FFN) is connected to each prediction from the decoder to predict the final class label, center coordinates, and dimensions of the bounding box. As in[31], the loss function contains Hungarian loss and bounding box loss.

**Jersey number recognition subsystem.** Jersey number recognition severely suffers from a data imbalance issue. Obtaining a balanced jersey number dataset with the jersey numbers from 1 to 99 requires a massive dataset for training. We propose to address this problem in two directions. First, rather than recognizing two-digit numbers, we strategically target single digit recognition, therefore dramatically reduce the demand for training data. Second, we employ RetinaNet[32,37] with focal loss design as the digit recognition step to alleviate the imbalance issue in the jersey number dataset.

With our single digit problem formulation, the number of classes is reduced from 100 to 10. With this significant reduction in the number of classes, it is then possible to achieve accurate results with a training dataset at the thousand level. Although the distribution of the 10 classes may still be uneven, the data imbalance issue will be further addressed by focal loss.

The jersey number recognition subsystem is applied to the detected players from the previous subsystem. Figure 5 shows the example of proposed detected players and the jersey numbers that are expected to be recognized. RetinaNet is a one-stage detection with fast performance due to a one-time process for each input image. Two essential components of the network are Feature Pyramid Networks (FPN) and Focal Loss. FPN is built on the top of ResNet-50, for the purpose of feature extraction, which therefore can take images of any arbitrary size as input and then output proportionally sized feature maps at multiple levels in the feature pyramid[38]. RetinaNet handles imbalance training classes better with the use of focal loss. Traditionally, class imbalance limits the performance of a classification model because most regions of an image can be classified as background and result in unhelpful contributions towards learning[39]. As a consequence, the gradient is overwhelmed and leads to degenerated models. Focal loss introduces a focusing parameter to cross entropy loss as defined by[40]:

$$FocalLoss = -\sum_{i=1}^{N} y_i log(p_i)(1-p_i)^{\gamma} \qquad (1)$$

where $N$ denotes the number of classes; $y_i$ is binary and equals 1 when $i$ belongs to a true label, otherwise is 0; $p_i$ is the probability distribution of the prediction of the $i$-th class; $\gamma$ is a focusing parameter to down-weight the loss of easily classified regions therefore differentiating the majority/minority regions.

**Game clock subsystem.** Game clock identification first extracts the game clock from the game video. Due to the rewinds and timeouts in the video, the exact timestamp of the game needs to be extracted from each frame. The subsystem then filters through the game clock data to derive individual play windows. It provides the information of the estimated start and end time of each play. The play windows feature the absolute range of frames for each play and are associated with the player detection subsystem to index the players present in each play.

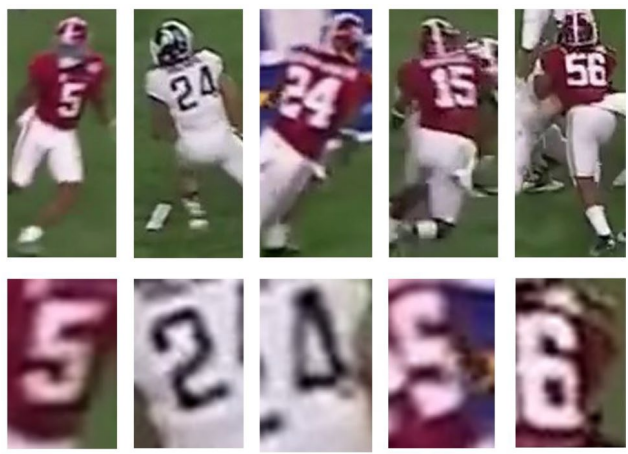

**Figure 5.** Example of players' proposals; example of digit on jersey.

*Detection of play clock information.* We choose a simple and fast solution, the Tesseract Optical Character Recognition (OCR)[41], to build the main software used for game clock subsystem because the clock region location in broadcast video has a distinctive background and uniform font as shown in lower left of Fig. 2. The program begins by first breaking down the video into individual frames. Individual frames are then cropped into specific regions to isolate the region with the game clock information. The cropped image is then altered three different ways to reach the desired OCR detection format. First, the image is grayscaled to enhance contrast between the text and background. Second, binary thresholding is applied to make the image black and white. Finally, the image is color-inverted to reach the desired OCR format, which is black text on a white background. After the preprocessing steps, the image is processed using the Tesseract OCR software for character detection and is outputted together with the absolute frame number. Zero is designated as output for scenes in which clock information is absent. As shown in Fig. 2, the information in the output text file is, from left to right, the absolute frame number, game clock information and play clock information. Based on this setup, the absolute frame index allows for simple synchronization with the player detection subsystem.

*Play window detection.* A play designation filter program is designed to sift through the text file and output play windows that contain the range of frames for each play. As shown in Fig. 2, the output of the play designation filter from left to right is play number, quarter number, frame at play start, frame at play end, play start time, and play end time.

**Database subsystem.** The database subsystem synchronizes the results from the game clock and player identification subsystems, and writes the results into a readable format. In addition to the input from the two subsystems, there are two additional inputs that we consider as prior knowledge: team jersey color and team roster.

Home and away team players are categorized by color filtering. Different features of a color histogram are used to differentiate teams. As examples shown in Fig. 6, in practice, the most representative region which is a small strip in the middle of player proposals is extracted to isolate the jersey. The *x* axis of the color histograms of images is the intensity from 0 to 255, and the *y* axis is the number of occurrences of that intensity. As an example of the red team player, the red channel intensity value is much higher than blue and green on average. Validation on a large number of red team jersey images analyzed in a similar manner agrees with the observation that the red intensity value is consistently much higher. On the other hand, a white jersey from the other team is hypothesized to show no dominant color channel. As shown in the corresponding color histogram, the white jersey has no dominant color channel. For the other jersey colors, a color filtering method can be easily generalized. Finally, jersey numbers are converted to the names of the players by roster.

## Experiments
**Implementation details.** In our experiments, we compare the performance of our player detection subsystem with a base line model: Faster R-CNN with ResNet50 backbone[34]. With each model, we empirically set the IoU to 0.75 to extract the detected bounding box. We compute the Average Precision (AP) and Average

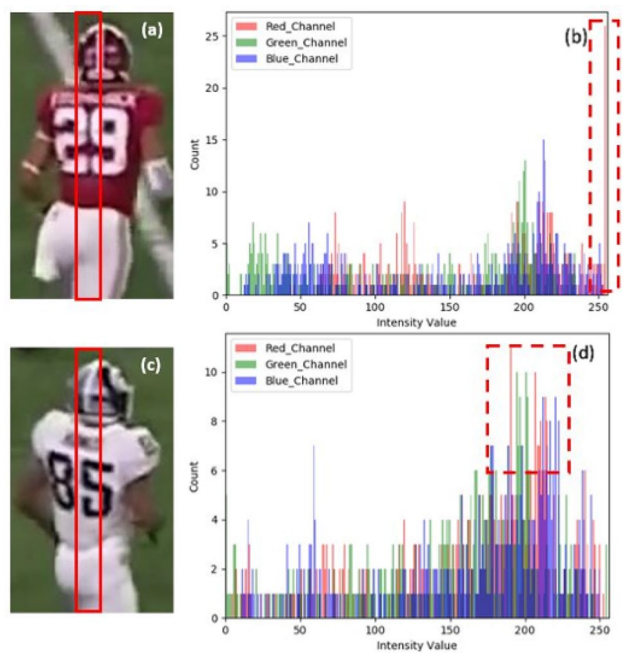

**Figure 6.** Color filtering demonstration: (**a**) player wears red jersey, (**b**) histogram of center region of red jersey, (**c**) player wears white jersey; (**d**) histogram of center region of white jersey.

Recall (AR) of *IoU* and different sized objects. The size of small object is from 32 × 32 to 96 × 96; the size of large object is above 96 × 96. We exclude the objects of size less than 32 × 32 because the jersey number is generally invisible when the whole player is included in such a small region. We transfer the weight that learned by[31] and further train the DETR with ResNet50 as backbone for 500 epochs until it converges using our football players dataset. For comparison, the pretrained Faster R-CNN with ResNet-50 backbone was trained on our dataset for 10,000 epochs until it converged. For each training epoch, we used a batch size of 2, learning rate of $10^{-4}$.

To evaluate the jersey number recognition subsystem, we trained the RetinaNet convolutional neural network with a pretrained weight from Model Zoo dataset[32]. All images are padded with zeros to avoid distortion. The model was trained for 20,000 epochs with a learning rate of $10^{-4}$ and a batch size of 81. We adopt the focal loss with focusing parameter $\gamma = 2$. The performance of this model is measured by assessing its mAP of detection and classification for all single digits. The optimal IoU threshold is found to be 0.55, and the optimal confidence threshold is 0.97. Note that this approach considers the two digits of a two-digit number independently. For example, the number "51" is labeled as a "5" and a "1", separately. Combining the digit recognition result on a left-to-right manner is further done as a post-processing step.

For the game clock subsystem, after clock information is extracted by the OCR, we set the condition for play switching as the period of 40 s of the game clock or the period of 5 s of the snap clock. We test the subsystem with game footage, and three evaluation criterion: the accuracy of the OCR, quarter transition and start/end time of each play. All experiments were carried out in parallel on four Quadro RTX 6000 GPUs.

**Dataset.** We validated our player detection subsystem using 10 highlight videos from Alabama Crimson Tide football game footages in season 2021–2022. As a proof of concept, we choose highlighted footage which are free of distractions caused by graphics and other media found in full-length footage. The videos are in standard high-definition (HD) display resolution of 1280 × 720 pixels. Frames captured from the video at 30 frames per second, 2866 of which are further filtered out for training and testing. To make sure the training and testing data are independent to each other, 2198 frames from 9 game videos are used for training; 668 frames from another game video are used for testing. For data augmentation, one copy of each image is given a motion-blurry effect. Therefore, the training set contains 4396 frames after augmentation, as examples shown in Fig. 4.

In the digit recognition subsystem, 4865 player proposals with visible jersey numbers from two Alabama Crimson Tide football game footages are manually labeled. Lastly, two copies of each image are created. One copy is a scaled version, and the other is a motion blurred version, which brings the final image count to 14,595, with 9730 of the images being augmented copies.

**Results and discussion.** *Player detection subsystem.* We quantitatively analyze the performance of the player detection subsystem. As shown in Table 1. We calculate the average precision of *IoU* from 0.5 to 0.95 with step size of 0.05. We perform both quantitative and qualitative comparison with Faster R-CNN in AP and AR of objects with small size and large size, respectively. The results show that DETR delivered better performance in an extremely crowded setting. With an IoU of 0.5, the average precision is 87.9%, which outperforms the Faster R-CNN by 8.9%. For an object with small and large size, the AP of the proposed model surpasses Faster R-CNN by 6.2% and 5.7%, and the AR of the proposed model are higher than the faster R-CNN by 7.7% and 1.9%, respectively.

To analyze the performance qualitatively, we list 3 representative images and their test results using different models. Figure 7a–c are original frames from a sparse scenario, moderate scenario, and crowded scenario. On the second row, faster R-CNN fails to detect the full range of the players proposal, therefore hurts the next digit recognition subsystem when only partials of jersey number are detected. The DETR in the last row outputs a desired result with the majority of players detected. Moreover, Faster R-CNN misses a lot of players in the crowded region. As the crowded scene is very common and is the focus of our work, we favor the result that the DETR delivered with almost every player fully detected. Both qualitative comparison in Fig. 7 and quantitative comparison in Table 1 support the superiority of DETR over Faster R-CNN.

To further validate the robustness of our propose framework, we test the player detection framework in other two football matches in a different season, 2022–2023: Alabama vs Texas A &M and Alabama vs Auburn. The precision, recall, and F1 scores are reported in Table 2. Each game is tested by 44 frames randomly sampled from the highlight video. We observe a consistently high precision, recall, and F1 among all three matches.

*Jersey number recognition subsystem.* The final break-down and distribution of the jersey number dataset is shown in Fig. 8a. The horizontal axis shows the individual digits, and the vertical axis shows the number of instances. All columns are colored to show the split between home and away jerseys. The digit 2 with the most occurrences is twice as frequent as the lowest digit 6, indicating a data imbalance issue generally existing in Jersey number dataset, even for a single-digit dataset. However, such imbalance issue is not dominant and we achieve a high mAP of 91.7% in evaluation.

In Fig. 8b, we show a confusion matrix which is normalized to the most popular class, digit 2. Digits 2,4,7,9 achieve higher accuracy compared to digits 6 and 8, which corresponds to the distribution of classes shown in Fig. 8a. Even though the utilization of focal loss alleviates the data imbalance issue to some extent, we notice that most of the incorrect classifications come from digits that were partially obscured or located on a player who was angled substantially with respect to the camera.

*Game clock subsystem.* Evaluation results are shown in Table 3. OCR performs well on recognizing the clock information with an accuracy of 95.6%. Play window detection achieves 100% accuracy on quarter transition, and an accuracy of 62% for start/end time designation.

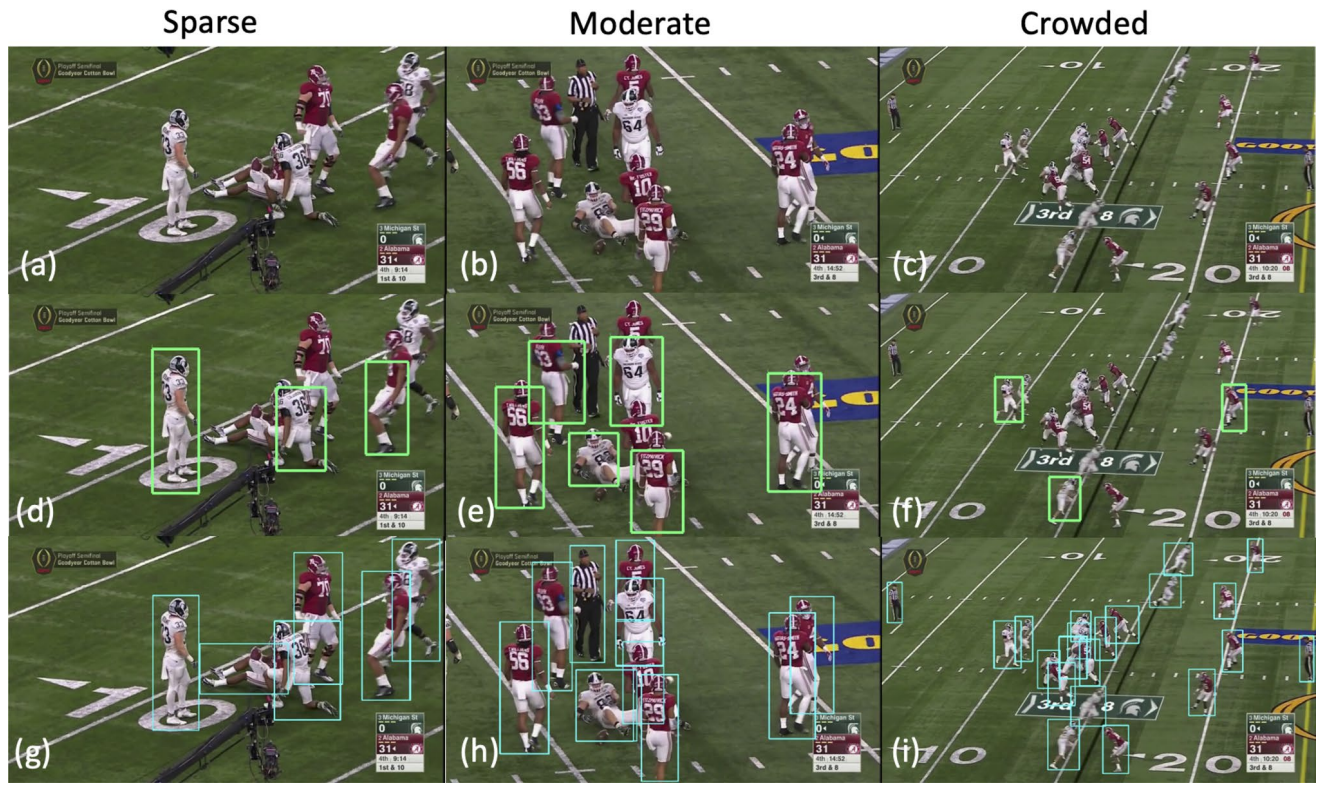

**Figure 7.** Representative testing results. (**a**–**c**) original frames; (**d**–**f**) Result from Faster R-CNN (**g**–**i**) Result from DETR.

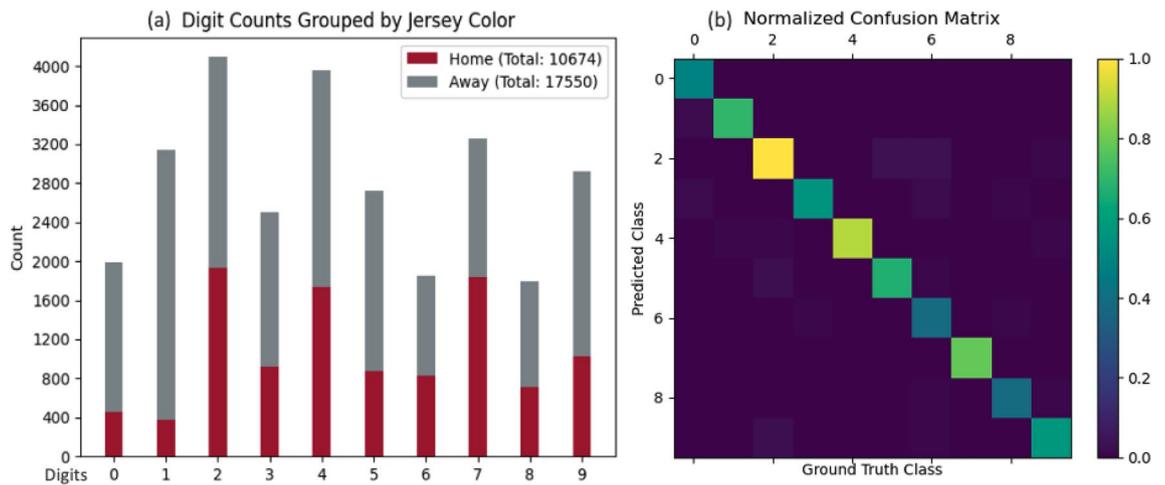

**Figure 8.** (**a**) Digit counts grouped by jersey color. (**b**) Confusion matrix of digit recognition result, with $IoU=0.55$ and confidence $= 0.97$.

In full game footage, the reset of the play clock can be used to track the end and subsequent start of each play. Although there are no standards for the end and start of a play in accelerated footage, it is possible to start a new play based on the jump in the play clock.

*Database subsystem.* We present a screenshot of data format in the integrated database in Fig. 9. Each play is related with game information, such as quarter number, start and end time, home and away team, and participating players of the home team. It is worth noting that the duplicated numbers are for offense and defense. However, it doesn't happen in the NFL because duplicated numbers are not allowed. A viewer could retrieve information and perform analysis either by play index or the jersey number of players. Integration of player information and clock information can alleviate the impact of players who may be partially occluded and invisible in some frames.

| | Faster R-CNN | DETR |
|---|---|---|
| $AP_{0.5:0.95}$ | 29.9 | 40.0 |
| $AP_{0.50}$ | 79.0 | 87.9 |
| $AP_{0.75}$ | 13.3 | 28.1 |
| $AP_{small}$ | 27.7 | 33.9 |
| $AP_{large}$ | 37.6 | 43.3 |
| $AR_{small}$ | 36.3 | 44.0 |
| $AR_{large}$ | 50.9 | 52.8 |

**Table 1.** Player detection performance. $AP_{0.5:0.95}$ denotes average precision of *IoU* from 0.5 to 0.95 with step size of 0.05. Average Precision (AP) and Average Recall (AR) of small (s) and large (l) objects are evaluated.

| Dataset | Precison | Recall | F1 |
|---|---|---|---|
| Alabama vs. Cincinnati (2021–2022) | 90.9 | 89.7 | 90.3 |
| Alabama vs. Texas A &M (2022–2023) | 77.8 | 97.8 | 86.7 |
| Alabama vs. Auburn (2022–2023) | 70.3 | 95.3 | 80.9 |

**Table 2.** Player detection performance from games.

| Evaluation | Criteria | Result |
|---|---|---|
| Clock detection | Time are read correctly | 95.6% |
| Quarter number | Quarter transitions are correct | 100% |
| Start/end time of each play | Play windows begin before the ball is snapped; end before the next snap of the ball | 62% |

**Table 3.** Evaluation of game clock subsystem.

| Play number | Quarter | Start time | End time | Home | Away | Participating players of Home team |
|---|---|---|---|---|---|---|
| 1 | 1 | 15:00 | 14:54 | Alabama | Michigan State | 3: Calvin Ridley or Bradley Sylve; 5: Ronnie Clark or Cyrus Jones; 15: Ronnie Harrison; 19: Reggie Ragland; 24: Geno Smith; 25: Dillon Lee; 26: Marlon Humphrey; 32: Rashaan Evans; 42: Keith Holcombe; 46: Michael Nysewander; 52: Andre Sims; 99: Adam Griffith |
| 2 | 1 | 13:53 | 13:24 | Alabama | Michigan State | 4: Daylon Charlot or Eddie Jackson; 5: Ronnie Clark or Cyrus Jones; 10: Reuben Foster; 15: Ronnie Harrison; 19: Reggie Ragland; 24: Geno Smith; 26: Marlon Humphrey; 29: Minkah Fitzpatrick; 86: A'Shawn Robinson; 93: Jonathan Allen |
| 3 | 1 | 12:41 | 12:00 | Alabama | Michigan State | 2: Tony Brown or Derrick Henry; 3: Calvin Ridley or Bradley Sylve; 13: ArDarius Stewart; 14: Jacob Coker; 70: Ryan Kelly; 74: Cam Rombinson; 76: Dominick Jackson; 84: Hale Hentges; 88: O.J. Howard |
| 4 | 1 | 11:37 | 11:18 | Alabama | Michigan State | 2: Tony Brown or Derrick Henry ArDarius Stewart; 14: Jacob Coker; 15: Ronnie Harrison; 50: Alphonse Taylor; 70: Ryan Kelly; 71: Ross Pierschbacher; 74: Cam Rombinson; 88: O.J. Howard |

······

**Figure 9.** Part of the output game log, including the player detecting information, play number, quarter, start/end time, home/away team.

### Discussion

In this paper, we propose an automatic analysis system for American football game based on deep learning. The experimental results demonstrate robustness in handling crowded scenarios and the capability to alleviate data imbalance issues. The proposed end-to-end system delivers complete detecting information using a single pass of football game footage. The uniquenesses of the proposed work are as follows. First, the proposed work gives a complete solution of player detecting and logging in the American football game, and provides expertise in handling the crowded setting of the football game. Currently, there is a lack of research on automatic analysis systems for American football game. Second, the proposed dataset uses bounding box labels solely, without other labeling requirement, such as human body labels. Therefore, the proposed system could be more generalizable with less labor needed.

For the future work, we plan to enhance the player detection by integrating information through consecutive frames and replenishing the players whose jersey numbers are not visible in some frames due to movement. The combination of spatial and temporal level feature is expected to be more sensitive to player's action[42]. The information on the players' positions could be incorporated into the jersey number recognition. For example, if the center is number 50, the wide receiver cannot also be number 50, so it may help to eliminate confusion between numbers. In addition, We will explore the possibility of identifying complicated human actions[43] from football video. Very small objects could be detectable by using ultra-high resolution camera or in combination with a super resolution method. Ball tracking could be fused to the current design and serves as complimentary information towards game clock logging, therefore helping with dividing the plays. The user interface that outputs the database will be further improved by developing a more robust graphic user interface. In addition, the framework is validated in highlight video in this study. To make the framework applicable to full match, we plan to integrate proposed framework with audio detection on whistle. We will use the occurrence of whistle as landmark to divide full match video into game-related contents, the later of which could be directly processed by proposed method.

## Conclusion

We demonstrate the feasibility of player detecting and logging in American football game. The proposed system enables player identification and time logging, so as to output the game log in a per-play basis. We propose a two-stage network design to first highlight player region and then identify jersey number in zoomed in sub-regions. Our player detection and jersey number recognition subsystems can be directly generalized to other football game footage. The qualitative and quantitative analysis have been thoroughly performed over three subsystems separately and over the entire system. The experimental result illustrates the reliability on handling the challenges of football game analysis.

## Data availibility

The datasets generated and analyzed in this work are available from the corresponding author upon reasonable request.

### Acknowledgements
The authors would like to thank Nguyen Hung Nguyen, Jeff Reidy, and Alex Ramey for their preliminary study, data annotation, and supportive experiment. Research reported in this paper was in part supported by NIH R21HD104164, NSF 2222739, and NSF 2239810.

### Author contributions
Hon.L.: conceptualization, methodology, formal analysis, validation, software, data curation, writing—original draft. N.W.: data curation, methodology, visualization. A.L.B., X. L.: data curation, writing—review and editing, validation. Hua.L: writing—review and editing. S.M.: conceptualization, writing—review and editing. Y.G.: conceptualization, methodology, supervision, writing—review and editing, funding acquisition.

### Competing interests
The authors declare no competing interests.

### Additional information
**Correspondence** and requests for materials should be addressed to Y.G.

**Reprints and permissions information** is available at www.nature.com/reprints.

**Publisher's note**  Springer Nature remains neutral with regard to jurisdictional claims in published maps and institutional affiliations.